\documentclass{article}
\usepackage{arxiv}
\usepackage[utf8]{inputenc} % allow utf-8 input
\usepackage[T1]{fontenc}    % use 8-bit T1 fonts
\usepackage{hyperref}       % hyperlinks
\usepackage{url}            % simple URL typesetting
\usepackage{booktabs}       % professional-quality tables
\usepackage{amsfonts}       % blackboard math symbols
\usepackage{nicefrac}       % compact symbols for 1/2, etc.
\usepackage{microtype}      % microtypography
\usepackage{lipsum}
\usepackage{fancyhdr}       % header
\usepackage{graphicx}       % graphics

\usepackage{amsmath}
\usepackage{mathtools}
\usepackage{amssymb}
\usepackage{bm}
\usepackage{caption}
\usepackage{subcaption}
\usepackage{float}
\usepackage{tikz}
\usepackage{algorithmic,algorithm}
\usepackage{import}
\usepackage{multirow}
\usepackage{enumitem}
\usepackage{tabularx}
\usepackage{xcolor}

\title{Multi-relation Message Passing for Multi-label Text Classification
\thanks{\textit{\underline{Citation}}: 
Ozmen, M. and Zhang, H. and Wang, P. and Coates, M. Multi-relation message passing for multi-label text classification. \textit{Proc. IEEE Int. Conf. Acoustics, Speech and Signal Processing (ICASSP)}, 2022.}
}

\author{
  Muberra Ozmen \\
  McGill University \\
  Montreal, Canada\\
  \texttt{muberra.ozmen@mail.mcgill.ca} \\
  \And
  Hao Zhang \\
  Hong Kong University of Science and Technology \\
  Hong Kong\\
  \texttt{hzhangcx@connect.ust.hk} \\
  \AND
  Pengyun Wang \\
  Huawei Noah’s Ark Lab \\
  Montreal, Canada\\
  \texttt{wangpengyun@huawei.com} \\
  \And
  Mark Coates \\
  McGill University \\
  Shenzhen, China\\
  \texttt{mark.coates@mcgill.ca} \\
}

\begin{document}

\maketitle

\begin{abstract}
A well-known challenge associated with the multi-label classification problem is modelling dependencies between labels. Most attempts at modelling label dependencies focus on co-occurrences, ignoring the valuable information that can be extracted by detecting label subsets that rarely occur together. For example, consider customer product reviews;  a product probably would not simultaneously be  tagged  by  both  ``recommended'' (i.e., reviewer is happy and recommends the product)  and ``urgent'' (i.e., the review suggests immediate action to remedy an unsatisfactory experience). Aside from the consideration of positive and negative dependencies, the direction of a relationship should also be considered. For a multi-label image classification problem, the ``ship'' and ``sea'' labels have an obvious dependency, but the presence of the former implies the latter much more strongly than the other way around. These examples motivate the modelling of multiple types of bi-directional relationships between labels. In this paper, we propose a novel method, entitled Multi-relation Message Passing (MrMP), for the multi-label classification problem. Experiments on benchmark multi-label text classification datasets show that the MrMP module yields similar or superior performance compared to state-of-the-art methods. The approach imposes only minor additional computational and memory overheads.
\end{abstract}


\section{Introduction}
Multi-label classification involves selecting the correct subset of tags for each instance from the available label set. There are numerous real world applications ranging from image annotation to document categorization~\cite{review_Madjarov2012}. A naive approach consists of converting the problem into multiple binary classification problems. This is the \emph{binary relevance (BR)} method~\cite{BR_Luaces2012}. A critical difference between multi-label and multi-class classification is that the class values are not mutually exclusive in multi-label learning. Usually, we anticipate that there are dependencies between the labels, and there are often multiple different types of relationships. Incorporating and learning the dependency relationships between labels during the learning process has an appealing potential of boosting predictive performance. 

The simplest approach that considers label dependencies is the \emph{label powerset (LP)} method \cite{review_Tsoumakas2007}. This involves converting the problem to multi-class classification by treating each possible combination of labels as a separate class, i.e., a multi-label problem with $L$ possible labels would be converted to a multi-class problem with $2^L$ classes. While the LP method enables us to make use of powerful multi-class classification methods, it suffers from scaling to problems with a large number of labels. Aside from the LP method, the existing methods that account for label dependencies construct the relationships between labels by assessing co-occurrences in the training data \cite{ML-CNN-GCN_ZhaoMin2019, LaMP_Lanchantin2019, MPVAE_Bai2020}. Co-occurrences represent a \emph{pulling} type of relationship between labels (labels that often appear together should be strongly encouraged to appear together in predictions). However, focusing solely on frequent co-occurrences ignores other valuable information. In particular, there should be a distinction between labels that occasionally appear together and those that {\em never} appear together. Aside from statistical label dependencies determined by co-occurrence (or the lack thereof), there are other valuable relations that should be considered when designing a classifier. These include semantic relations (e.g., synonyms, plural forms) or pre-defined structural relations (e.g., hierarchical category relations). 

In this paper, we propose a novel method, called \emph{Multi-relation Message Passing (MrMP)}, which employs a \emph{Compositional Graph Convolutional Network (CompGCN)} ~\cite{CompGCN_Vashishth2020} to model multiple bi-directional relationships between labels. We make the following contributions:
\begin{enumerate}
    \item We design a multi-relation label embedding module that can be integrated into most encoder-decoder type neural network architectures to account for the dependencies between labels. 
    \item We propose a simple and efficient method to extract statistically significant \textit{pulling} and \textit{pushing} relations between labels.
    \item We design a \emph{Transformer} \cite{transformer_Waswani2017} based message passing network for the multi-label text classification problem considering two types of statistical relations between labels.
    \item We perform experiments on benchmark multi-label text classification datasets, comparing with classical and state-of-the-art baselines, to demonstrate the efficacy of our method.
\end{enumerate}

\section{Related Work}
There have been multiple attempts in the literature to capture label dependencies for multi-label classification problems. \emph{Probabilistic classifier chains (PCC)} stack a sequence of binary classifiers and predict one label at a time conditioned on previously predicted labels \cite{CC_Read2011, PCC_Dembczynski2010, Rectifying-CC_Senge2019}. PCC based methods decompose the joint probability of observing label subsets into a product of conditional probabilities. The computational complexity therefore increases exponentially with the number of possible labels. To reduce the length of the classifier chain and the corresponding model complexity, \cite{ML-RNN_Nam2017} suggests \emph{ML-RNN}, a sequence-to-sequence architecture that focuses on predicting positive labels only. PCC models can suffer from the sub-optimal pre-defined label ordering. Training and inference time is an issue due to the inability to harness parallel computation.

In latent embedding learning methods the inputs and outputs are projected into a shared latent space \cite{extMLC_Bhati2015, 2stage_LE_Chen2019, MC-GM_Ma2020}. An effective recent method is the \emph{Multivariate Probit Variational AutoEncoder (MPVAE)}~\cite{MPVAE_Bai2020}. This maps features and labels to probabilistic subspaces and uses the Multivariate Probit (MP) model to make predictions by sampling from these subspaces. A shared covariance matrix is learned to capture label dependencies. However, the embedded learning methods fail to incorporate prior knowledge about label structures. 

Graph-based methods for capturing label dependencies have shown promising performance \cite{ML-CNN-GCN_ZhaoMin2019, KSSNet_Wang2019}. \cite{ML-CNN-GCN_ZhaoMin2019} use \emph{Graph Convolutional Networks (GCNs)} \cite{GCN_Kipf2017} to map label representations to inter-dependent object classifiers for the multi-label image classification task. In these methods, a label graph is typically built based on label co-occurrence, with nodes corresponding to labels and edges corresponding to how two labels interact. The label graph can be combined with graph neural networks to enable the interactive learning of features and label embeddings. \cite{LaMP_Lanchantin2019} and \cite{HOTVAE_zhao2021} propose starting with a simple fully-connected graph or a co-occurrence label graph. A \emph{Message Passing Neural Network (MPNN)} then passes messages among label embeddings and features to enable learning of high order label correlation structure that is conditioned on the features. Due to the greater flexibility offered by MPNNs compared to GCNs, these models achieve state-of-the-art performance on many benchmark datasets. However, all of the above graph methods use simple undirected graphs, neglecting the richness of information present in the joint empirical probability distribution of the labels, such as whether the occurrence of label A encourages or suppresses the occurrence of label B, relative to the marginal distribution of the latter.  

This work proposes the construction of a multi-relational label graph and uses multi-relational graph neural networks to more comprehensively capture the information present in the joint empirical probability distribution of the labels. The inclusion of the richer information in the model can potentially improve performance for multi-label classification problems.

\section{Multi-relation Message Passing (MrMP)}
\begin{figure*}[t]
	\centering
	\includegraphics[width=0.9\textwidth]{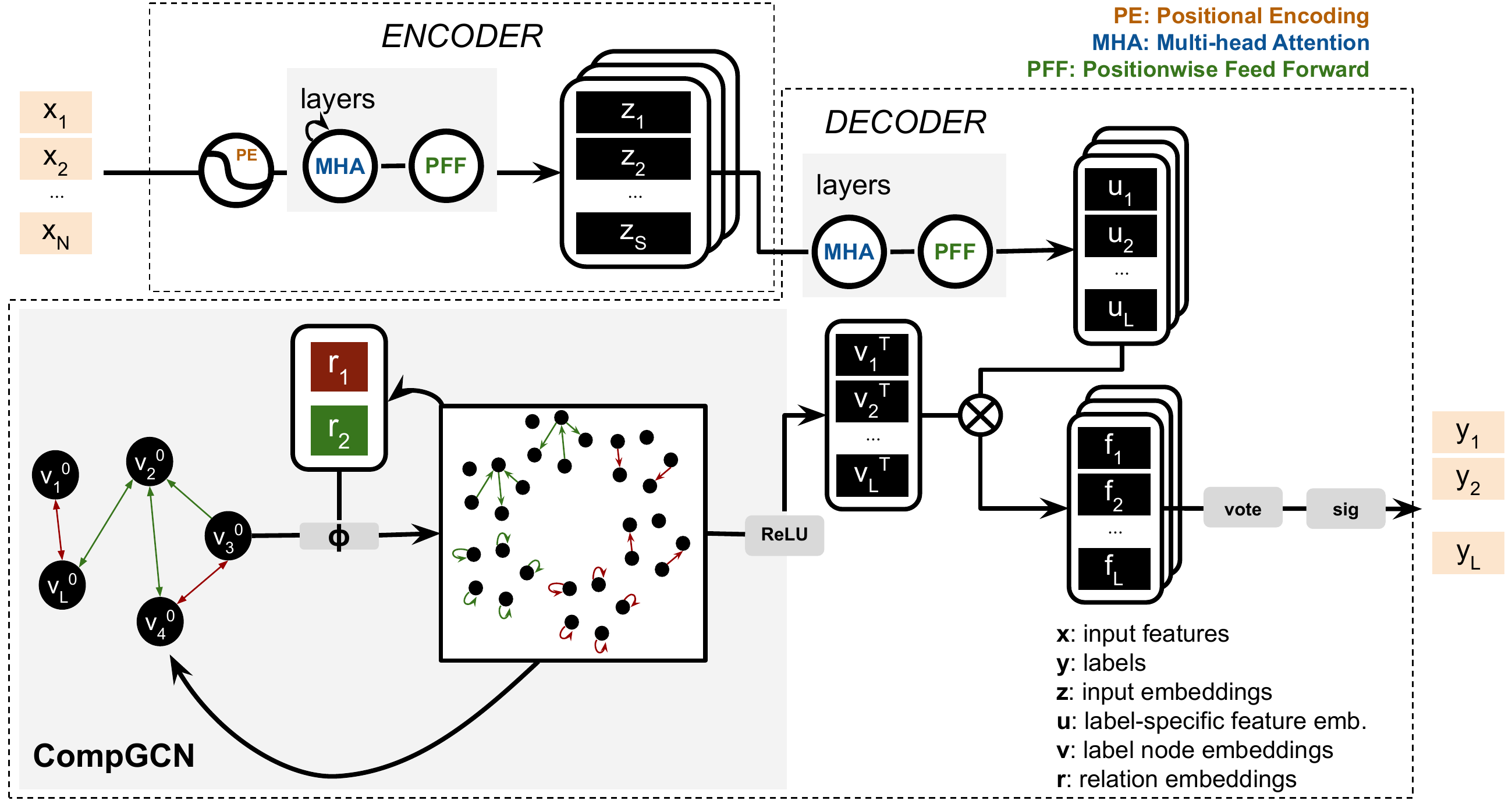}
	\caption{Overview of MrMP. The encoder first applies positional embedding on the input sequence $\bm{x}$. The output then passes through encoder-self attention layers of MHA and a PFF to obtain the input word embeddings $\bm{z}$. The decoder initializes label $\bm{v^0}$ and relation embeddings $\bm{r}$ and composes them together using a function $\phi$ to prepare the multi-relation layers' input. The layers of the multi-relation module aggregate the label embeddings, updated by \textbf{\textcolor{green}{pulling}} and \textbf{\textcolor{red}{pushing}} relation sets, and compose them with updated relation embeddings recursively to feed the next layer. The label node embeddings $\bm{v^T}$ are obtained by applying a ReLU on the final layer's output. The encoder-decoder attention layers of MHA and PFF prepare label-specific feature embeddings $\bm{u}$. $\bm{f}$ is obtained by element-wise product of $\bm{v}$ and $\bm{u}$. The label predictions $\bm{\hat{y}}$ are obtained after voting on $\bm{f}$ caused by multiple layers of encoder output and applying a sigmoid to normalize.}
	\label{fig:model}
\end{figure*}

For the multi-label classification problem, the advantage of using Message Passing Neural Networks (MPNNs) arises due to the structural power of graphs for representing global label dependencies. The use of RNN based architectures has two often overlooked advantages. The ordering in the prediction path allows the architecture to focus on the next most probable label and ignore irrelevant labels. Moreover, the ordering models dependencies in a directed manner, albeit in only one direction. Our proposed method strives to combine these advantages. The proposed model contains an encoder for extracting features and a decoder for predicting labels. Although the method can be applied to a variety of types of input datasets, we focus on the input type of text to provide a more concrete description.

\subsection{Notation and Preliminaries}
We denote a sample by a sequence of words $\bm{x}$ which is composed of $N$ components $\bm{x}=(x_1, x_2, ..., x_N)$ where $N$ is the length of sequence. Associated with each sample is its corresponding label set, represented as a binary vector $\bm{y}=(y_1, y_2, \dots , y_L)$, where $y_i\in \{0,1\}$ indicates whether label $i$ appears. We use a training set $(\bm{x}_j,\bm{y}_j)_{j\in \mathcal{T}}$ to learn the parameters of a predictive model.  
The model then takes $\bm{x}$ as an input and outputs $\bm{\hat{y}}=\{\hat{y}_1, \hat{y}_2, ..., \hat{y}_L\}$ where $0\leq \hat{y}_i\leq 1$ is the estimated probability that label $i$ appears. We denote input feature embeddings by $\bm{z} = \{z_1, ..., z_S\}$ and label specific feature embeddings by $\bm{u} = \{u_1, ..., u_L\}$. The estimations are evaluated by mean binary cross entropy $\mathcal{L}_{\text {bce}}(\bm{y},\bm{\hat{y}}) = \frac{1}{L} \sum_{i=1}^{L} - y_{i} \log (\hat{y}_{i}) + (1-y_{i}) \log (1-\hat{y}_{i})$. \\

\noindent\textbf{Multi-head Attention (MHA)} is performed by concatenating parallel heads of scaled dot-product attention. The main motivation is to prevent destabilization during training by bad initialization of learnable weights. The inputs are generalized as $\bm{Q}$ for ``Query", $\bm{K}$ for ``Keys", and $\bm{V}$ for ``Value".  Scaled-dot product attention is formulated as follows:
\begin{equation}
    \text{Attention}(\bm{Q},\bm{K},\bm{V}) = \text{softmax}\bigg(\frac{\bm{Q} \bm{K}^T}{\sqrt{d_k}}\bigg)\bm{V}\,.
\end{equation}

\noindent\textbf{Positionwise Feed Forward (PFF)} networks are fully connected feed-forward network layers which consist of two linear transformations, with a ReLU activation between them:
\begin{equation}
    \text{PFF}(\bm{x}) = \max(0, \bm{x} \bm{W}_1 + b_1)\bm{W}_2 +b_2\,.
\end{equation}

\subsection{Proposed Model}
We consider two types of statistical relations between labels; \emph{pulling} and \emph{pushing}. Let a relation graph be represented by adjacency matrices $\bm{A}^+$ and $\bm{A}^-$ for pulling and pushing edges, respectively. The method used to calculate the corresponding adjacency matrices will be explained later. To properly utilize the multi-relational adjacency matrices obtained, we propose a Multi-relation Message Passing (MrMP) method. An overview of the model is provided in Figure \ref{fig:model}. In essence, it  combines  the  ideas  of  LaMP \cite{LaMP_Lanchantin2019} and CompGCN~\cite{CompGCN_Vashishth2020}. Assume that the input is a sequence of tokens $\bm{x}=\{x_1, x_2, ..., x_N\}$. The encoder initializes input word embeddings and applies positional encoding if the data type is sequential \cite{transformer_Waswani2017}, then stacks several layers of self-attention to construct the sequential representation of input at each layer, $\bm{z}^l=\{z_1^l, z_2^l, ...,z_S^l\}$. The decoder is composed of a label-relation module and a label-feature module. The label-relation module is responsible for capturing the multi-relational dependencies among labels, while the label-feature module extracts the most relevant features for each label. 

The label-relation module initializes the label node embeddings and stacks CompGCN layers with pulling and pushing relations. Let $z_+$ and $z_-$ represent pulling and pushing relation embeddings, respectively. Let $\bm{W_+}$ and $\bm{W_-}$ be trainable weight matrices for pulling and pushing relations, respectively. At each layer of the label relation module the label embeddings $\bm{V}^{l+1} = [v_1^{l+1}, v_2^{l+1}, ..., v_L^{l+1}]$ are updated based on the output from the previous layer $\bm{V}^{l}$ as follows:
\begin{equation}
    v_i^{l+1} = f\bigg(\sum_{j \in \mathcal{N}^+(i)} \bm{W}_+^l \phi(v_i^l, z_+^l) + \sum_{j \in \mathcal{N}^-(i)} \bm{W}_-^l \phi(v_i^l, z_-^l) \bigg)\,.
\end{equation}
Here the neighbourhoods for each relation, $\mathcal{N}^+(i)$ and $\mathcal{N}^-(i)$, include the original relations, inverse relations and self loops. In our experiments we set $\phi(\cdot,\cdot)$, the composition function that combines the label hidden state and the relation hidden state, to be summation. In our case, since we also have an intuitive understanding of the relationship between the two relations,  we set the relation embeddings to be $z_+^l = -z_-^l$ at each layer. The relation embeddings are updated at each layer by relation specific trainable parameters, $z_+^{l+1} = \bm{W}_{rel}z_+^l$.

We use the final label embeddings $\bm{V}^T$ as queries to extract label-specific features $\bm{U}$ based on the output from the encoder. Since different labels may be most effectively determined by features at different layers, we extract label specific features $\bm{U}^l$ based on each $\bm{z}^l$, respectively. On a given decoder layer the label specific feature embeddings are updated as follows:
\begin{align}
    m_i^l &= u_i^l+\sum_{j=1}^{S} \text{MHA}(u_i^l, z_j^l ; \bm{W}_r^l), \\
    u_i^{l+1} &= m_i^l + \text{PFF}(m_i^l; \bm{W}_r)\,,
\end{align}
where $u_i^0=v_i^T$. Then, label prediction is performed based on each $\bm{U}^l$ to produce $\hat{\bm{f}}^l = \text{diag}(\bm{U}^l*\bm{V}^T)$ where $*$ stands for element-wise multiplication. Finally, the ultimate prediction for class probabilities $\hat{\textbf{y}}$ is determined by a voting, $\hat{\bm{y}} = \sum_{l} \hat{\bm{f}}^l$. \\

\noindent\textbf{Calculating Label Relation Graphs.} The conventional approach for calculation of the prior graph that represents the relationship between labels involves estimating conditional probabilities based on co-occurrences in the training data. LaMP (with prior)~\cite{LaMP_Lanchantin2019} uses a simpler method that forms an edge between two labels if they co-occur in any training sample. We consider two types of statistical relations between labels: pulling and pushing. We first test the existence of a relation based on occurrences for each label pair, and then decide whether that relation is pushing or pulling. That is, given the presence of dependency between two labels, we hypothesize whether it corresponds to co-occurrence or avoidance relation. For all label pairs $i$ and $j$,
\begin{enumerate}
    \item Perform a dependence test by setting null hypothesis as $H_0: P(L_j = 1 | L_i = 1) \neq P(L_j = 1)$. We test this via a chi-squared test statistic on a pairwise contingency table.
    \item Determine type of relation: If the label pair passes the hypothesis test, the existence of label $j$ significantly depends on the existence of label $i$. The label pair is set to have a pulling edge if $P(L_j = 1 | L_i = 1) > P(L_j = 1)$, and a pushing edge otherwise. 
\end{enumerate}

\noindent\textbf{Relational Loss Function.} We use a combination of cross-entropy loss and a relation based label embedding distance for training our model. Let the final label node embedding for label $i$ be denoted by $\hat{v}_i$. The relation based label embedding distance is formulated as follows: 
\begin{equation}
        \mathcal{L}_{\text {rel}}(\bm{\hat{v}}_i) = - \frac{1}{|\mathcal{N}^+(i)|}\sum_{j \in \mathcal{N}^+(i)} \frac{\bm{\hat{v}}_i . \bm{\hat{v}}_j}{\lVert \bm{\hat{v}}_i \rVert \lVert \bm{\hat{v}}_j \rVert} + \frac{1}{|\mathcal{N}^-(i)|}\sum_{j \in \mathcal{N}^-(i)} \frac{\bm{\hat{v}}_i . \bm{\hat{v}}_j}{\lVert \bm{\hat{v}}_i \rVert \lVert\bm{\hat{v}}_j \rVert}\,.
    \end{equation}

\section{Experiments}

\subsection{Datasets and Evaluation Metrics}
Although the proposed method can be extended to different domains, currently we have experimented with 4 benchmark multi-label text classification datasets.  Bibtex, Bookmarks \cite{bibtex_bookmarks_Katakis08} and Delicious \cite{delicious_Tsoumakas2008} involve automated tag suggestion for entries from the BibSonomy social publication, the Bookmark sharing system, and for web pages from the del.ico.us social bookmarking site, respectively. Reuters-21578 \cite{reuters_Lewis2004} is a collection of newswire stories. Our experiments cover a range of dataset sizes between 7,538 and 804,410. The number of labels range between 90 and 983; and vocabulary size ranges between 500 and 50,000. We use the same data splits as described in~\cite{LaMP_Lanchantin2019} for fair comparison. The dataset details are provided in Table \ref{tab:dataset_statistics}

\begin{table*}[t]
\caption{Dataset Statistics}
\centering
\begin{tabular}{@{}ccccccccc@{}}
\toprule
Dataset & Input Type & \#Instances & \#Labels & \#Features & Label Cardinality & \#Train & \#Valid & \#Test \\ \midrule
Bibtex & Binary vector & 7,538 & 159 & 1,836 & 2.38 & 4,377 & 777 & 3,019 \\
Bookmarks & Binary vector & 87,856 & 208 & 2,154 & 2.03 & 48,000 & 12,000 & 27,856 \\
Delicious & Binary vector & 16,071 & 983 & 500 & 19.06 & 11,597 & 1,289 & 3,185 \\
Reuters-21578 & Sequential & 10,789 & 90 & 23,662 & 1.23 & 6,993 & 487 & 2,515 \\
\bottomrule
\end{tabular}
\label{tab:dataset_statistics}
\end{table*}

The instance-based performance metrics we use to evaluate our method are example based F1 score (ebF1), and subset accuracy (ACC). The label-based performance metrics are micro-averaged F1 score (miF1), and macro-averaged F1 score (maF1). Subset accuracy measures the fraction of times that an algorithm identifies the correct subset of labels for each instance. The example-based F1 score is aggregated over samples and the macro-averaged F1 score over labels. The micro-averaged F1 score takes the average of the F1 score weighted by the contribution of each label, and thus takes label imbalance into account. 

\begin{itemize}
    \item Instance based: Let there are $L$ number of labels and $M$ number of samples, $\bm{y}_i = (y_{i1}, ..., y_{iL})$ and $\hat{\bm{y}}_i = (\hat{y}_{i1}, ..., \hat{y}_{iL})$ denote binary vector of ground-truth and predicted labels on sample $i$ respectively.
        \begin{itemize}
            \item \textbf{ACC} $ = \frac{1}{M} \sum_{i=1}^M I[\bm{y}_i = \hat{\bm{y}}_i]$
            \item \textbf{ebF1} $ = \frac{1}{M} \sum_{i=1}^{M} \dfrac{2 \sum_{j=1}^L y_{ij} \hat{y}_{ij}}{\sum_{j=1}^L y_{ij} + \sum_{j=1}^L \hat{y}_{ij}}$ 
        \end{itemize}{}
    \item Label based: each label $y_j$ is treated as a separate binary classification problem (each label has its own confusion matrix of true-positives (tp), false-positives (fp), true-negatives (tn), false-negatives (fn).)
        \begin{itemize}
            \item \textbf{miF1} $ = \dfrac{\sum_{j=1}^L 2tp_j}{\sum_{j=1}^L 2tp_j +fp_j +fn_j}$
            \item \textbf{maF1} $ = \dfrac{1}{L} \sum_{j=1}^L \frac{2tp_j}{2tp_j +fp_j +fn_j}$
        \end{itemize}{}
\end{itemize}

\subsection{Baselines}
In addition to state-of-art methods ML-RNN \cite{ML-RNN_Nam2017}, LaMP \cite{LaMP_Lanchantin2019}, and MPVAE \cite{MPVAE_Bai2020}, we compare our algorithm to baseline classifiers such as ML-KNN \cite{mlknn_Zhang2007} and ML-ARAM \cite{mlaram_Benites2015}. ML-KNN finds the nearest examples to a test class by the k-Nearest Neighbors algorithm and selects assigned labels using Bayesian inference. ML-ARAM use Adaptive Resonance Theory (ART) based clustering and Bayesian inference to calculate label probabilities. The results for ML-KNN and ML-ARAM are obtained by functions provided in the scikit-learn library \cite{scikit-learn}. For ML-RNN and LaMP we provide the results reported in the corresponding papers. For MPVAE we have reproduced the results in order to obtain performance metrics for the datasets under study. We provide the reported result if it exists and is better than those we reproduced.\footnote{The reproduced results of MPVAE are in italic font in the tables.} 

\subsection{Implementation Details}
We follow \cite{LaMP_Lanchantin2019} for setting the relevant hyperparameters. For all datasets, the latent model dimensionality of the neural network is set to 512, the number of encoder and decoder layers to 2, and the number of attention heads to 4. The dropout probability is 0.2 for Bibtex and Reuters, and 0.1 for the other datasets. The significance level of the dependency test to calculate label graphs' adjacency matrices is selected to be 0.05. The value of the threshold used to convert soft prediction to predicted class is determined using the validation sets and optimized individually for all performance metrics. The model is trained with a batch size of 32, and the Adam \cite{adam_kingma2017} optimizer is used to compute gradients and update parameters with an initial learning rate of 0.0002 and step size 10 for 10\% decay rate. 

\subsection{Results}
\begin{table*}[t]
    \caption{Experiment Results}
    \centering
    \small
    \import{./}{"results"}
    \label{tab:results}
\end{table*}

The results are provided in Table \ref{tab:results}. Overall our experiments show that Multi-relation Message Passing (MrMP) yields better performance than the state-of-art methods most of the time. The value of the multi-relation approach is most evident in subset accuracy in which MrMP outperforms the baselines by \%7 on average. The overall performance improvements for the  ebF1, miF1 and maF1 metrics are also positive, but below 2\%. The  maF1 improvement is slightly greater than the miF1 improvement. Comparatively, maF1 focuses more on rare labels; this is in line with the intuition that the guidance provided by incorporating multiple relations is more useful for labels with low observation frequency.

\noindent\textbf{ML-KNN and ML-ARAM.} MrMP yields better results than ML-KNN and ML-ARAM for all metrics since these two algorithms rely only on clustering of input word embeddings for mapping features to labels, while MrMP employs a message passing neural network where input word embeddings are mapped to label embeddings via multi-head attention. 

\noindent\textbf{ML-RNN.} ML-RNN has the capability of generating input embeddings by taking the sequential nature into account. By following a prediction path, the algorithm is able to focus on the next probable label given a previously predicted set of labels. This enhances the capability of capturing the full label subset for each instance, so ML-RNN is most competitive for the subset accuracy metric. The results on Reuters-21578 show that MrMP outperforms ML-RNN by 5\% and 14\% in miF1 and maF1 while achieving slightly better subset accuracy. The integration of the MrMP module with pulling and pushing relations allows the architecture to emulate the advantage of making predictions sequentially, while enjoying the flexibility and computational efficiency of an MPNN. 

\noindent\textbf{LaMP.} Comparison to LaMP reveals the true impact of the multi-relation approach in terms of capturing label dependencies. While our model is most similar to LaMP in terms of the main building blocks of mapping features to labels, we employ multi-relation GCNs to model label dependencies instead of decoder self attention as in LaMP. The enhancement of subset accuracy performance compared to LaMP reveals the importance of accounting for the pushing type of relation between labels.  

\noindent\textbf{MPVAE.} The comparisons with MPVAE show that multi-head attention based mapping of input to output and multi-relation based label dependency modelling are more effective than using a multivariate probit model for the mapping and a global covariance matrix for dependency modelling. 

\subsection{Ablation}
\begin{figure*}
	\centering
	\includegraphics[trim=0 40 0 40, clip, width=0.8\textwidth]{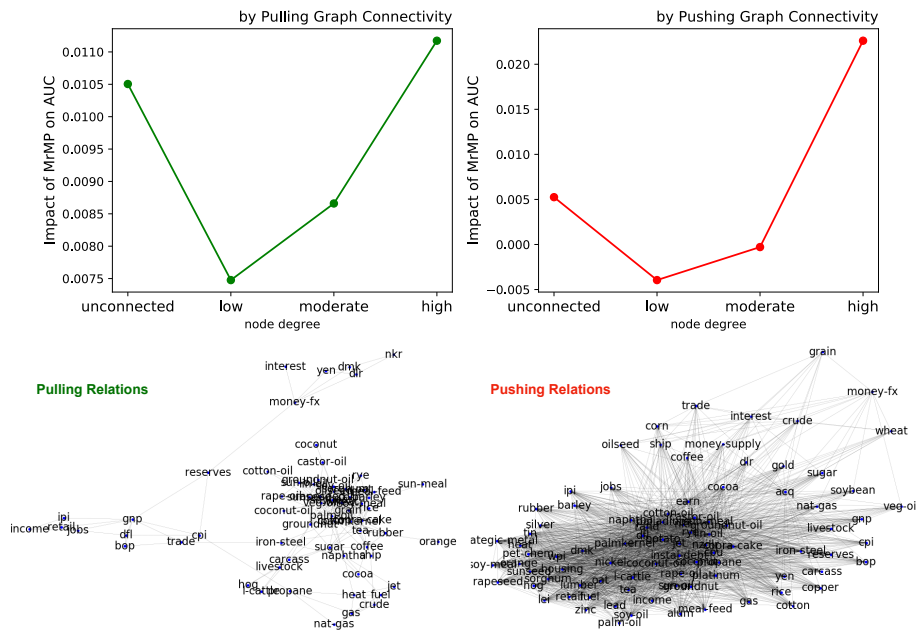}
	\caption{Ablation study on Reuters-21578 dataset: First, the node degree of labels in each relation graph shown in figure are calculated by summing the number of edges and the labels are grouped into four. Second, the differences of AUC between model with and without MrMP are calculated per label. Third, for each label group differences on AUC score are aggregated by taking average. Between the connected label nodes (have at least 1 edge) It can be observed that the impact of MrMP module increases by the node degree in pulling and pushing relation graphs. This shows that the strength of MrMP emerges from modelling complex label dependencies better.}
	\label{fig:cs}
\end{figure*}

In order to examine the effectiveness of multi-relation module, we perform an ablation study by assessing the performance of the proposed architecture with and without the MrMP module. Figure \ref{fig:cs} shows the results of this analysis for Reuters-21578 dataset as an example \footnote{AUC$_j$ for a label $j$ is calculated by by the area under the Receiver Operating Characteristic (ROC) curve which is plotted by calculating true positive and and false positive rate for a set of soft prediction conversion thresholds \cite{review_Madjarov2012}.}.

\subsection{Complexity}
Let $N$ represent maximum sequence length, $L$ represent number of labels, $R$ represent number of relations, $d$ and $d'$ represent hidden dimension and inner model dimension respectively (which we set $d'=2 \times d$ in our experiments). For encoder we have encoder self attention of $\mathcal{O}(N^2 \cdot d)$ and positionwise feed forward of $\mathcal{O}(N \cdot d \cdot d')$ per layer, which reaches to $\mathcal{O}(N \cdot d \cdot (N+d'))$. For decoder we have encoder-decoder attention of $\mathcal{O}(N \cdot L \cdot d)$, positionwise feed forward of $\mathcal{O}(L \cdot d \cdot d')$, and relational module of $\mathcal{O}(d^2 + r \cdot d + r^2)$ per layer. In our experimental setting the overall complexity could be simplified to $\mathcal{O}((N+L) \cdot d^2 + (N^2 + N \cdot L) \cdot d)$. 

\section{Conclusion}
In this paper, we propose a Multi-relation Message Passing Network (MrMP) for the multi-label classification problem. The proposed model can be adapted to any number of relations as long as the prior is known. We consider \textit{pulling} and \textit{pushing} statistical relations between labels and use the Composition-based GCN to learn label embeddings that reflect these statistical relations.  The experiments show that MrMP outperforms state-of-art models in terms of effectively modelling label dependencies in order to achieve improved classification performance \footnote{The implementation of our method is publicly available at \texttt{https://github.com/muberraozmen/MrMP}}.

\bibliographystyle{unsrt}

\end{document}